%% file: main.tex
\pdfoutput=1

\documentclass[11pt]{article}

\input{preamble}

\usepackage{algorithm}
\usepackage{algpseudocode}
\algnewcommand{\LeftComment}[1]{\Statex #1}

\usepackage[final]{acl} 

\usepackage{times}
\usepackage{latexsym}

\usepackage[T1]{fontenc}

\usepackage[utf8]{inputenc}

\usepackage{microtype}

\usepackage{inconsolata}

\usepackage{graphicx}

\newif\ifcomment
\commenttrue
\ifcomment
\newcommand{\hh}[1]{{\bf \textcolor{red}{HH: #1}}}
\else
\newcommand{\hh}[1]{}
\fi

\title{Demystifying optimized prompts in language models}

\author{
  Rimon Melamed $^{1}$
  Lucas H. McCabe $^{1,2}$
  H. Howie Huang $^{1}$
  \\ \\
$^1$~The George Washington University \\
$^2$~LMI Consulting\\
\texttt{\{rmelamed,lucasmccabe,howie\}@gwu.edu}
}

\begin{document}
\maketitle
\begin{abstract}
    Modern language models (LMs) are not robust to out-of-distribution inputs. Machine generated (``optimized'') prompts can be used to modulate LM outputs and induce specific behaviors while appearing completely uninterpretable. In this work, we investigate the composition of optimized prompts, as well as the mechanisms by which LMs parse and build predictions from optimized prompts. We find that optimized prompts primarily consist of punctuation and noun tokens which are more rare in the training data. Internally, optimized prompts are clearly distinguishable from natural language counterparts based on sparse subsets of the model's activations. Across various families of instruction-tuned models, optimized prompts follow a similar path in how their representations form through the network. %
    ~\footnote{Code and models available at \url{https://github.com/rimon15/demyst_optim_prompts}}
\end{abstract}

\section{Introduction}

Language models (LMs)~\citep{grattafiori2024llama3herdmodels, pmlr-v202-biderman23a, gemmateam2024gemmaopenmodelsbased, abdin2024phi3technicalreporthighly} are trained on large amounts of filtered internet data~\citep{gao2020pile800gbdatasetdiverse, JMLR:v21:20-074, penedo2024finewebdatasetsdecantingweb, soldaini2024dolmaopencorpustrillion}, which consist primarily of interpretable natural language text. Recent work has found that these models are sensitive to machine-generated \textit{optimized prompts}, which, although seemingly uninterpretable, can be used to elicit targeted behaviors~\citep{shin-etal-2020-autoprompt, wen2023hard, zou2023universaltransferableadversarialattacks, melamed-etal-2024-prompts}. Specifically, we define optimized prompts as prompts that are generated via the gradient-based discrete prompt optimization method called Greedy Coordinate Gradient (GCG)~\citep{zou2023universaltransferableadversarialattacks}; see Section~\ref{sec:related} for further background.

In this work, we seek to better understand the underlying mechanisms by which language models parse these seemingly garbled inputs. In particular, we ask the question:

\begin{center}
    \textit{Are discretely optimized prompts truly uninterpretable?}
\end{center}

This question has major implications in several areas, including safety and privacy. Specifically, discrete prompt optimization has commonly been applied in the adversarial setting to ``jailbreak'' LMs, resulting in toxic or undesirable behavior~\citep{zou2023universaltransferableadversarialattacks,liao2024amplegcg, andriushchenko2024jailbreakingleadingsafetyalignedllms,zhu2024autodan}; see Section~\ref{sec:related} for further details. A better understanding of these optimized prompts is crucial to ensure robustness and safety in LMs.

To this end, we explore the nature of optimized prompts through experiments which consider both the discrete makeup of optimized prompts, as well as how these prompts are processed internally by LMs; see Section~\ref{sec:exp}.

\begin{table*}[t]
    \centering
    \caption{Examples of prompt pairs and their most influential tokens.
        For each token we show its text and influence score (higher means a larger behavioral change when removed). Both natural language prompts and optimized prompts rely on punctuation, which typically appears at the end of the prompts. This can be attributed to the auto-regressive nature of the models, where the final token can have a pronounced influence; see Section~\ref{sec:important-toks}.}
    \small
    \begin{tabular}{p{0.23\textwidth}|p{0.23\textwidth}|p{0.23\textwidth}|p{0.23\textwidth}}
        \toprule
        \textbf{Original prompt}                                                                                                     & \textbf{Top-3 original removals} & \textbf{Optimized prompt} & \textbf{Top-3 optimized removals} \\
        \midrule
        \multicolumn{4}{c}{\textbf{word-stories}}                                                                                                                                                                                       \\
        \midrule
        Tom was nice and they played together in the grass .                                                                         &
        \begin{minipage}[t]{0.23\textwidth}
            . (\textbf{14.80})

            Tom (\textbf{7.65})

            played (\textbf{2.13})
        \end{minipage}                                                                                          &
        Bee Squeak Dickie paw Angeles Wee Table Bananas goat Jazz Tom least care Or raking pinched waved glanced dancers .           &
        \begin{minipage}[t]{0.23\textwidth}
            . (\textbf{16.23})

            Bee (\textbf{2.75})

            dancers (\textbf{0.74})
        \end{minipage}                                                                                                                                                                                              \\
        \midrule
        Bob proudly showed them the picture he had just printed .                                                                    &
        \begin{minipage}[t]{0.23\textwidth}
            . (\textbf{17.82})

            Bob (\textbf{1.07})

            picture (\textbf{0.94})
        \end{minipage}                                                                                          &
        Timothy telling None display colours page wipes Pete visor beamed Their recognise 3 deleted pear symbol mitten show puzzle ! &
        \begin{minipage}[t]{0.23\textwidth}
            ! (\textbf{14.04})

            Timothy (\textbf{2.55})

            Pete (\textbf{0.70})
        \end{minipage}                                                                                                                                                                                              \\
        \midrule
        \multicolumn{4}{c}{\textbf{Pythia-1.4B}}                                                                                                                                                                                        \\
        \midrule
        Construct a web address for a book recommendation website.                                                                   &
        \begin{minipage}[t]{0.23\textwidth}
            . (\textbf{4.00})

            Construct (\textbf{2.04})

            book (\textbf{1.74})
        \end{minipage}                                                                                          &
        onas books auored A gateway URL:** EzAzureongeOm orn Yorker OKnote?).                                                        &
        \begin{minipage}[t]{0.23\textwidth}
            ?). (\textbf{3.68})

            URL (\textbf{2.65})

            :** (\textbf{0.77})
        \end{minipage}                                                                                                                                                                                              \\
        \midrule
        At around 10:30 a.m.,                                                                                                        &
        \begin{minipage}[t]{0.23\textwidth}
            ., (\textbf{7.83})

            m (\textbf{1.48})

            At (\textbf{1.02})
        \end{minipage}                                                                                          &
        irling Singh Dillonanchez approached approached detectives HertEDem CLEC=\{traceSONumbled 700 EVENT).\$),                    &
        \begin{minipage}[t]{0.23\textwidth}
            \$), (\textbf{2.78})

            detectives (\textbf{1.27})

            ). (\textbf{1.24})
        \end{minipage}                                                                                                                                                                                              \\
        \bottomrule
    \end{tabular}
    \label{tab:token-removal-examples}
\end{table*}

\subsection{Our contributions}
To the best of our knowledge, this is the first work which systematically investigates optimized prompts over a wide range of models.

\paragraph{Optimized prompts consist of influential and specific tokens.}

We find that both natural language and optimized prompts consist of specific ``influential'' tokens which have an out-sized impact on eliciting desired behavior, and these influential tokens consist primarily of nouns and punctuation; see Table~\ref{tab:token-removal-examples} for examples of these prompts and Section~\ref{sec:important-toks} for details.

\paragraph{Optimized prompts rely on rare tokens.}

When comparing tokens in both optimized and natural language prompts to the pre-training corpus, we find that the majority of tokens in optimized prompts are \textit{more rare} with respect to the training data than their natural language counterparts. Furthermore, the token distribution of optimized prompts visibly deviates from standard Zipfian behavior; see Section~\ref{sec:rarity}.

\paragraph{Optimized prompts have distinct internal representations.}

We train sparse probing classifiers to distinguish between optimized and natural language prompts based on their activations, and find that these classifiers achieve high accuracy even under sparsity constraints.
These findings suggest fundamental differences in how optimized prompts and natural language prompts are represented internally; see Section~\ref{sec:internal-rep}.

\section{Related work}\label{sec:related}

\paragraph{Discrete prompt optimization} Discrete optimization for prompt-based LMs typically consists of perturbing a set of arbitrary tokens in a meaningful way in order to induce desired behavior. Pioneering work includes HotFlip~\citep{ebrahimi-etal-2018-hotflip} which finds adversarial examples for character-level neural classifiers by performing guided token substitutions based on gradient information. AutoPrompt~\citep{shin-etal-2020-autoprompt} builds on the HotFlip algorithm, and appends ``trigger'' tokens to the prompts of masked language models such as BERT~\citep{devlin-etal-2019-bert}. These trigger tokens are modified in a similar fashion to HotFlip, and are used to improve performance on downstream tasks such as sentiment analysis and natural language inference (NLI). More recently, ~\citealp{zou2023universaltransferableadversarialattacks} introduce Greedy Coordinate Gradient (GCG), which uses an algorithm similar to AutoPrompt to find adversarial triggers which elicit desired output in modern decoder LMs.

Modern LMs undergo an alignment process~\citep{NEURIPS2022_b1efde53, rafailov2023direct} which is meant to improve model safety and refusal to harmful instructions~\citep{bai2022constitutionalaiharmlessnessai}. Typically, the goal of discrete optimization is to  ``jailbreak'' these models, and cause them to operate outside of their aligned state~\citep{zou2023universaltransferableadversarialattacks, zhu2024autodan, liao2024amplegcg, guo2024coldattack, thompson2024flrtfluentstudentteacherredteaming, andriushchenko2024jailbreakingleadingsafetyalignedllms}, resulting in malicious output and degraded performance on downstream tasks.

\paragraph{Language model interpretability}
Several prior works attempt to shed light on the black-box nature of neural language models. \Citealp{elhage2021mathematical} take a mechanistic circuit-based approach, examining how individual neurons and connections impact model predictions. They view the model's outputs at each layer as the ``residual stream'', a communication channel that each individual layer can modify.

In contrast, other work adopts a high level view by examining model outputs at a representation level
~\citep{zou2023representationengineeringtopdownapproach,wu2024reft} through various means such as linear probes~\citep{alain2017understanding, gurnee2023finding} and sparse autoencoders~\citep{bricken2023monosemanticity, huben2024sparse}. Several works explore the dynamics by which LMs promote concepts and representations. \Citealp{logitlens} investigates how predictions are built by projecting each of the model layer's outputs to the vocabulary space. \Citealp{geva-etal-2021-transformer} find that transformer feed-forward layers serve as key-value memories, and encode interpretable concepts and patterns. Furthermore, LM predictions appear to be constructed by propogating representations that are interpretable in the vocabulary space~\citep{geva-etal-2022-transformer, belrose2023elicitinglatentpredictionstransformers}. In our work, we apply several techniques such as sparse probing and projections to the vocabulary space in order to study how LMs build predictions for optimized prompts.

\paragraph{Analyzing machine generated prompts} There have been several investigations probing the properties of discretely optimized prompts. \Citealp{ishibashi-etal-2023-evaluating} explore the robustness of prompts optimized via AutoPrompt~\citep{shin-etal-2020-autoprompt}, and find that these prompts are highly sensitive to token ordering and removal when evaluated on NLI tasks. Similarly \Citealp{cherepanova2024talking} find that GCG optimized prompts can be degraded via token-level perturbations. Furthermore, machine generated prompts are easier to generate if the target text is shorter and comes from an in-distribution dataset such as Wikipedia~\citep{cherepanova2024talking}. In contrast, \Citealp{kervadec2023unnatural} examine the attention patterns and activations of optimized prompts for two OPT models~\citep{zhang2022optopenpretrainedtransformer}, finding that optimized prompt tend to trigger distinct ``pathways'' in the model, which differ from how natural language prompts are processed.

Concurrent work~\citep{rakotonirina2024eviltwinsevilqualitative} explores properties of prompts optimized via GCG, and find that these prompts consist of several ``filler'' tokens which do not affect the generation, and that the effectiveness of these prompts relies heavily on the last token. They also discover that there exist local dependencies within gibberish prompts based on specific keywords and bigrams. On the other hand, in our work we investigate the properties of optimized prompts both from a token perspective by training a new model with a word-level tokenizer, as well as from the perspective of the model's internal representations via hidden state analysis, probing, and causal intervention.

\section{Experimental setup}\label{sec:exp}
We focus our work on transformer decoder~\citep{NIPS2017_3f5ee243,Radford2018ImprovingLU} models. We use the Tiny Stories~\cite{eldan2023tinystoriessmalllanguagemodels} dataset, which consists of synthetically generated stories meant to be understandable by a three year old child. We train a transformer decoder language model based on the GPT-NeoX~\cite{black-etal-2022-gpt, pmlr-v202-biderman23a} architecture; see Appendix~\ref{app:stories_setup} for full training details. Originally, the model uses a Byte-pair encoding (BPE) tokenizer~\citep{sennrich-etal-2016-neural}, which results in optimized prompts that include several nonsensical characters and subwords~\cite{melamed-etal-2024-prompts, cherepanova2024talking}. Because we wish to better understand which specific words appear in optimized prompts, we train a new word-level tokenizer over the Tiny Stories corpus. Using word-level tokenization allows us to better interpret optimized prompts, since we do not need to extrapolate meaning from sub-word tokens and can directly evaluate each word in the prompt individually.

\begin{figure}[t]
    \centering
    \includegraphics[width=0.48\textwidth]{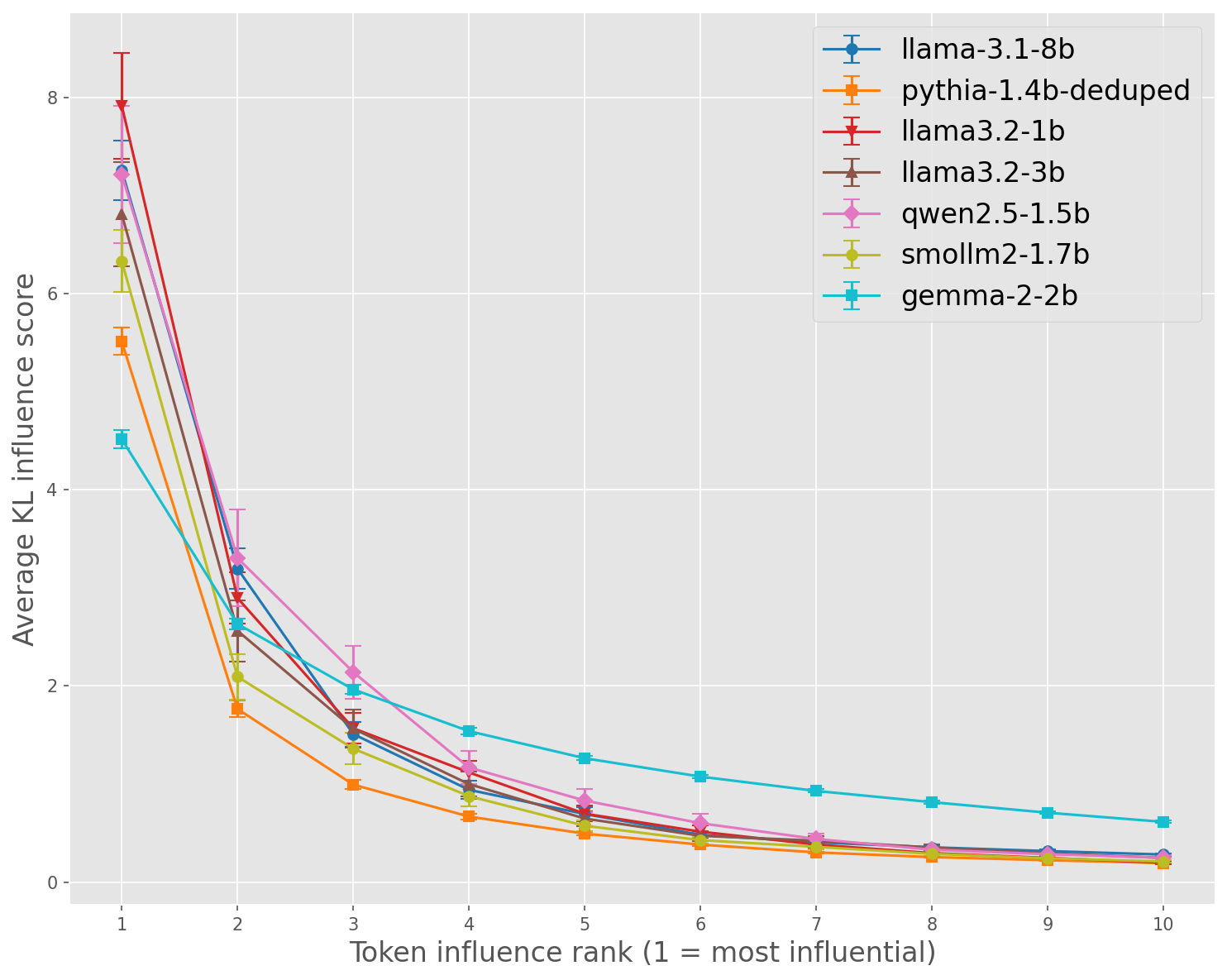}
    \caption{Token rank influence. The influence score is computed via Equation~\ref{eq:influence_score}.
        We find that the most influential token for each prompt has an out-sized effect.}
    \label{fig:tok_influence}
\end{figure}

\begin{figure*}
    \centering
    \includegraphics[width=1.0\linewidth]{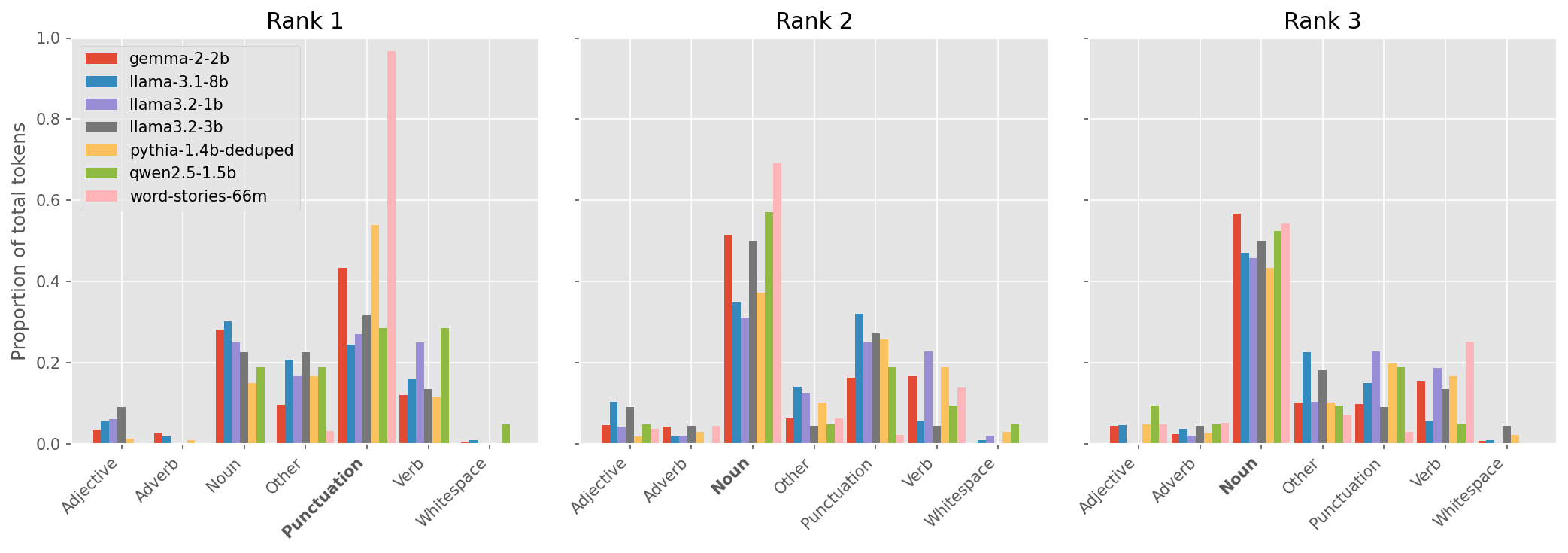}
    \caption{Token category analysis by rank. For each model and token influence
        rank (as computed in Section~\ref{sec:important-toks}), we show the proportion of tokens belonging to each part of speech category. The most common category at each rank is highlighted in bold.  While the specific distributions vary between models, nouns consistently make up the largest portion of tokens (with the exception of rank 1 in the base models, where punctuation dominates).}
    \label{fig:grammar_graph}
\end{figure*}

In addition to the word-level Tiny Stories model, we optimize prompts using 18 open models from various model families, including both base models, and their instruction-tuned variants which have been aligned for chat purposes.
We use a variety of datasets for the optimization including Alpaca~\citep{alpaca}, WikiText-103~\citep{wikitext}, OpenHermes-2.5~\citep{OpenHermes-2.5}, and Dolly-15k~\citep{DatabricksBlog2023DollyV2}; see Appendix~\ref{app:data_setup} for further details.

In order to perform the discrete optimization, we use the ``evil twins'' framework~\citep{melamed-etal-2024-prompts}.
Formally, given a natural language prompt $\bp^* \in \R^{k \times V}$ which is a sequence of $k$ tokens mapped to the LM's vocabulary $V$, the objective is to find a new prompt $\bp \in \R^{l \times V}$ with $l$ tokens which is \textit{functionally} similar to $\bp^*$. This optimization corresponds to an empirical approximation of the KL divergence between $\bp^*$ and $\bp$, and is realized by sampling a set of continuations from the LM, $\bd_1,\ldots,\bd_n \sim \P_{\mathrm{LM}}(\cdot|\bp^*)$, and running the Greedy Coordinate Gradient (GCG) algorithm~\citep{zou2023universaltransferableadversarialattacks}. For the full algorithm and further details we refer the reader to Appendix~\ref{app:gcg} and~\Citealp{melamed-etal-2024-prompts}.

The KL divergence between prompts is defined as

\begin{multline}\label{eq:kl}
    d_{KL}(\bp^* || \bp) = \frac{1}{n} \sum_{i=1}^{n} \log(\P_{\mathrm{LM}}(\bd_i | \bp^*)) \\
    - \log(\P_{\mathrm{LM}}(\bd_i | \bp)).
\end{multline}

The lower $d_{KL}(\bp^* || \bp)$ is, the more functionally similar $\bp^*$ and $\bp$ are, and $d_{KL}(\bp^* || \bp) = 0 $ if and only if the two prompts are functionally equivalent~\citep{melamed-etal-2024-prompts}.

\section{Optimized prompts consist of specific influential tokens}\label{sec:important-toks}

Using the set of optimized prompts from the LMs, we analyze the influence of each token in the prompt by removing each token and measuring the change in KL divergence to the prompt with the token kept. Specifically, given an optimized prompt $\bp = [p_1, ..., p_k]$ consisting of $k$ tokens, we define the influence score $s_i$ of token $i$ as

\begin{align}\label{eq:influence_score}
    s_i = d_{KL}(\bp || \bp_{-i}),
\end{align}

where $\bp_{-i} = [p_1, ..., p_{i-1}, p_{i+1}, ..., p_k]$ is the prompt with token $i$ removed, and $d_{KL}$ is defined in Equation~\ref{eq:kl}. A larger influence score indicates that removing token $i$ causes a greater deviation from the functional behavior of the original prompt.

For each optimized prompt, we sort its tokens by their influence scores in descending order to obtain token ranks, where rank 1 corresponds to the most influential token (highest influence score $s_i$). We group tokens from all optimized prompts by their rank in order to understand their composition at different influence levels. We find that the most influential token (rank 1) has an outsized effect, and tokens at higher ranks have minimal influence; see Figure~\ref{fig:tok_influence}. Natural language prompts follow a similar pattern, which we describe in Appendix~\ref{app:orig_removals}. These findings are consistent with recent work indicating that optimized prompts largely consist of ``filler'' tokens that minimally impact prompt behavior~\citep{rakotonirina2024eviltwinsevilqualitative}.

\begin{figure}[ht]
    \centering
    \includegraphics[width=0.45\textwidth]{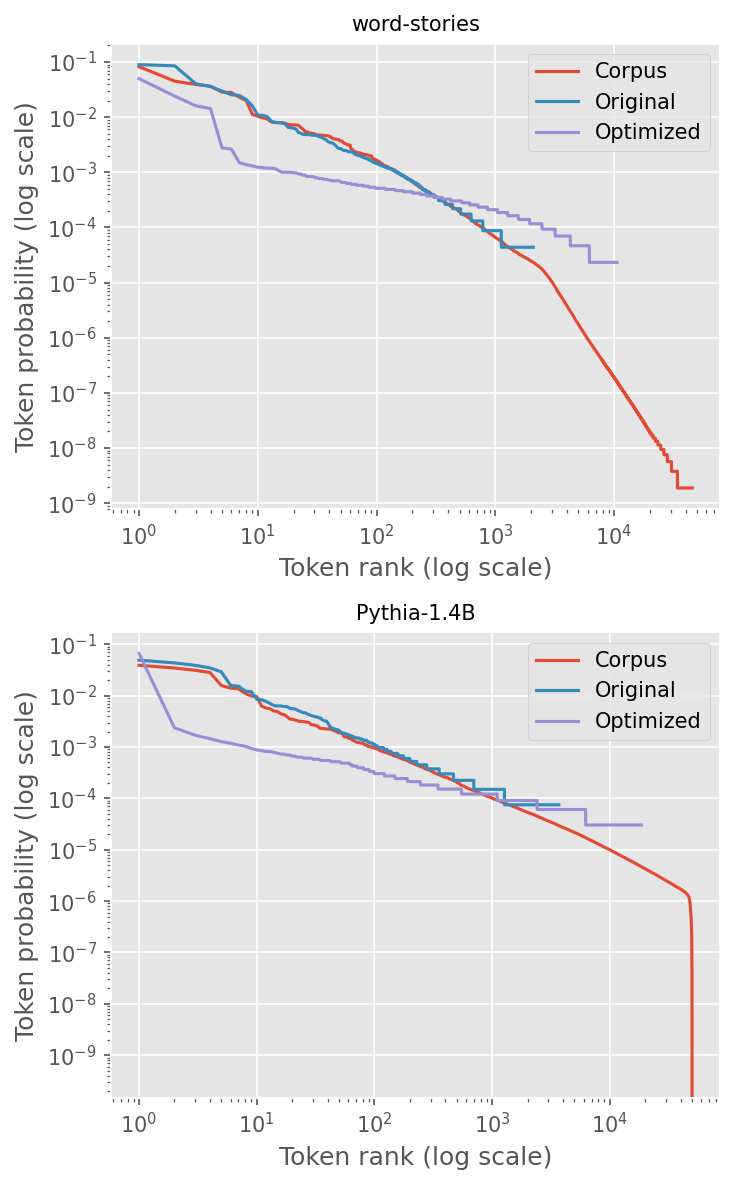}
    \caption{Zipf plots of token frequencies (excluding the end-of-sequence token) in the corpus, original prompts, and optimized prompts. The token distribution of optimized prompts visibly deviates from the expected Zipfian behavior.
    }
    \label{fig:tok_distributions}
\end{figure}

\subsection{Grammatical categories of optimized prompt tokens}

Given that certain tokens have an outsized impact on the prompt, we explore the grammatical makeup of these influential tokens. We perform part-of-speech tagging on each token in each prompt using spaCy~\citep{spacy2}. Interestingly, we find that punctuation forms the largest proportion of most influential (rank-1) tokens.
In addition, for all models, nouns consistently make up the largest portion of tokens; see Table~\ref{tab:token-removal-examples} for examples of these prompts and Figure~\ref{fig:grammar_graph} for full results. Furthermore, these trends are not unique to optimized prompts, as natural language prompts are also dependent on punctuation and nouns; see Appendix~\ref{app:orig_removals}.

\section{Optimized prompts use rare tokens}\label{sec:rarity}

For the word-stories and Pythia-1.4b models where we have access to the pre-training corpus, we further analyze the frequency of tokens in both natural language prompts and optimized prompts.

\begin{figure}[t]
    \centering
    \includegraphics[width=0.45\textwidth]{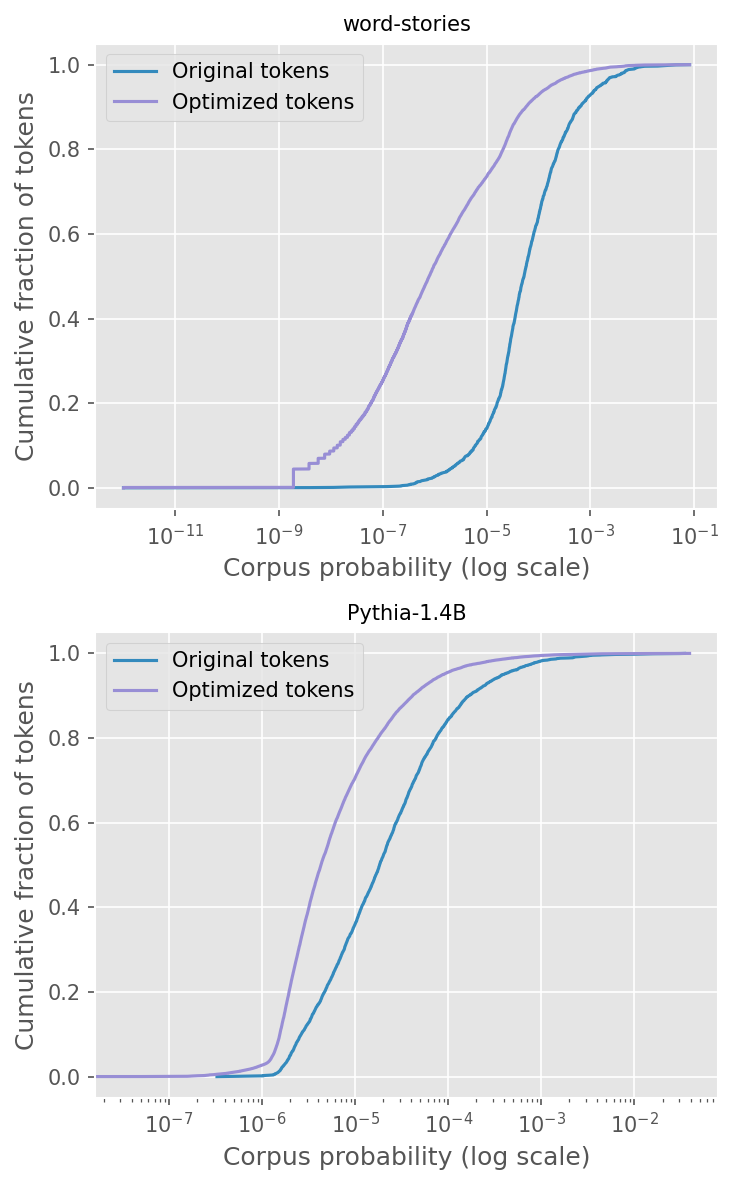}
    \caption{
        CDF of token corpus-frequency. For each token used by either the original natural language or the optimized prompts, we plot its probability of appearing in the training corpus, versus the cumulative fraction of tokens up to that probability. The optimized prompts rely more on corpus-rare tokens than their original natural language counterparts.
    }
    \label{fig:relative_token_frequency}
\end{figure}

\subsection{Optimized prompts do not look like natural language (distributionally)}

The distribution of tokens in both the corpus and original prompts exhibit power law-like behavior, consistent with the Zipfian distribution of natural language. In contrast, the sub-linear behavior for optimized prompts in log-transformed space indicates that there are fewer tokens with high frequencies than a power law would predict (Figure \ref{fig:tok_distributions}). This is underscored by normalized entropy (i.e., entropy divided by that of a uniform distribution over the same alphabet size), which is much higher for the optimized prompts' token distribution (0.8968 for word-stories, 0.9338 for Pythia) vs. that of the original prompts (0.7102 for word-stories, 0.7988 for Pythia).

\begin{figure*}[ht]
    \centering
    \includegraphics[width=1.0\textwidth]{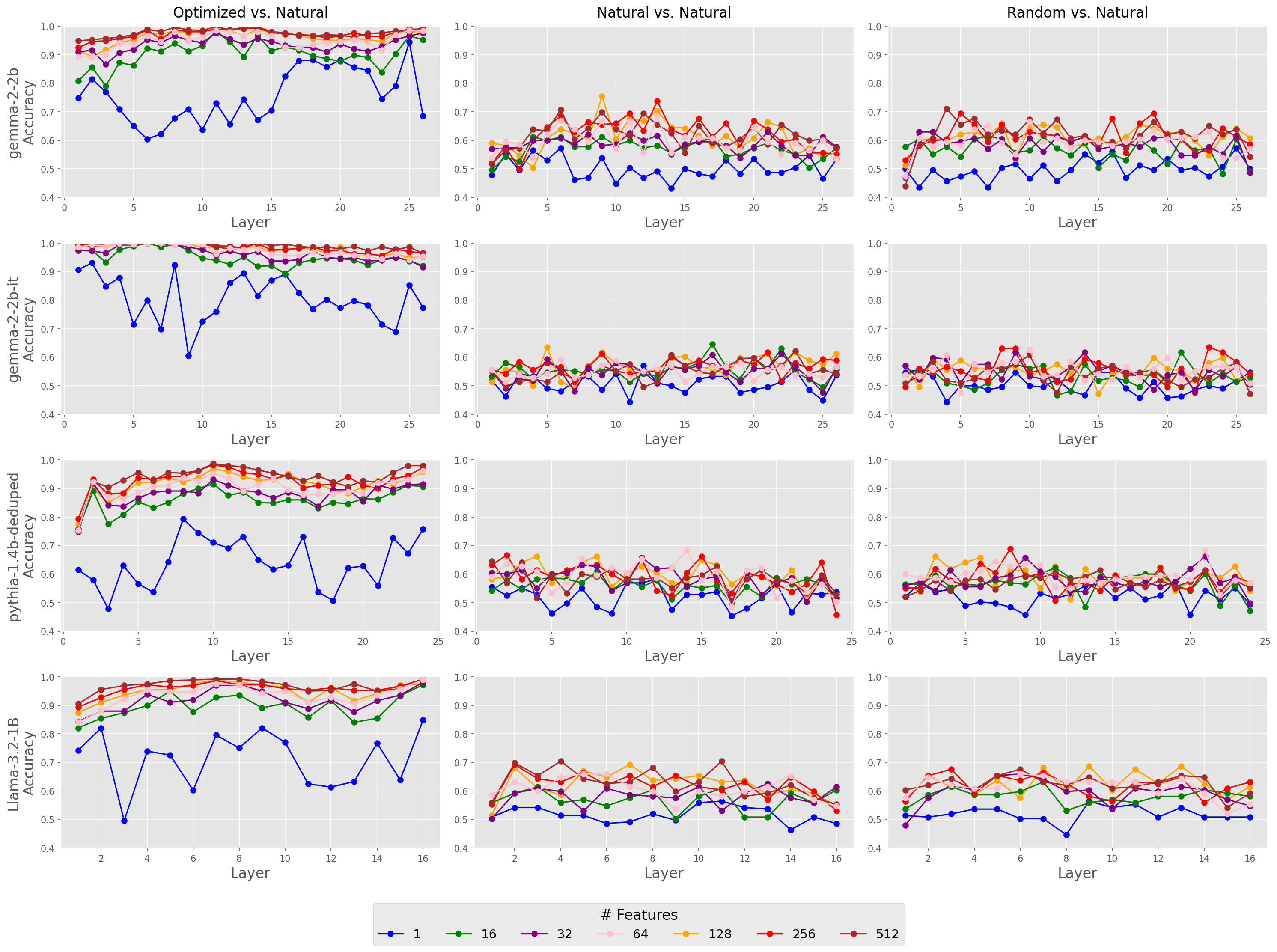}
    \caption{Sparse linear classifier probe accuracy on top-$k$ features of the model output at each layer. We identify top features using Equation~\ref{eq:mean-diff}, and train a linear classifier to discriminate between optimized and natural language prompts. The first column compares optimized and natural language prompts. The second and third columns show a baseline comparison of natural language prompts vs. other natural language prompts, and of natural language prompts vs. random prompts, respectively. We refer the reader to Figure~\ref{fig:app_probe_results_all} in Appendix~\ref{app:full_res} for the full results on all models.
    }
    \label{fig:probe-acc}
\end{figure*}

\subsection{Optimized prompts rely on tokens that are rare in the training data}

For convenience, we use the shorthand \textit{corpus-rare} to refer to tokens that are rare in the training data and \textit{corpus-common} for those that are common in the training data. Natural language prompts tend to use more corpus-common tokens than their optimized counterparts; see Figure~\ref{fig:relative_token_frequency}. Prior work finds that LMs are sensitive to tokens which are under-trained and not found as frequently in the training corpus, dubbed ``glitch tokens''~\citep{solidgoldmagikarp,li2024glitchtokenslargelanguage,land-bartolo-2024-fishing}.
The higher frequency of \textit{corpus-rare} tokens may be due to the fact that these tokens are potentially under-trained, and are thus more likely to have a stronger signal during the optimization procedure.

\begin{figure*}[ht]
    \centering
    \includegraphics[width=1.0\textwidth]{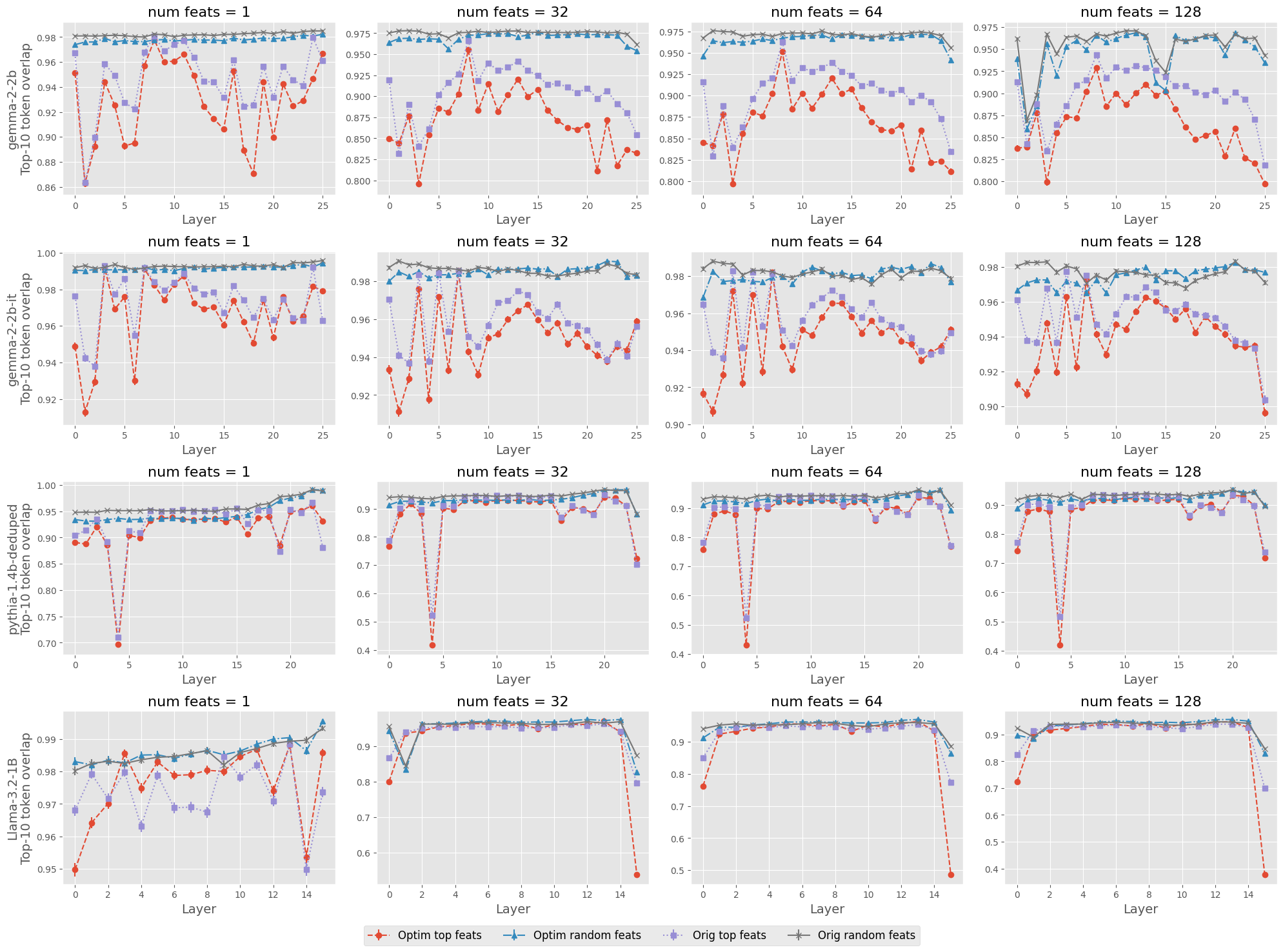}
    \caption{Average top-10 token prediction overlap. The overlap is computed via Equation~\ref{eq:overlap}.
        Overall, the importance of the top identified features appears to be model dependent, and it is not necessarily the case that the optimized prompts rely more heavily on these features. However, it is clear that there are certain layers which influence the output more (specifically the first and last layers); see Figure~\ref{fig:app_intervene_results_all} in Appendix~\ref{app:full_res} for the full results.
    }
    \label{fig:top-intervened}
\end{figure*}

\section{Internal representations of optimized prompts}\label{sec:internal-rep}

Given that optimized prompts consist of corpus-rare tokens and differ significantly in composition from natural language prompts, we ask whether the same differences exist within the model's internal representations.

\subsection{Sparse probing for optimized prompts}\label{sec:probe}
We investigate whether it is possible to detect optimized prompts purely from the model's activations. Given optimized and natural language prompt pairs, we follow~\Citealp{gurnee2023finding} and train sparse probe classifiers at each layer to differentiate optimized prompts from their natural language counterparts. Concretely, each transformer block in the network consists of a multi-head self-attention layer (MHSA) and a multi-layer perceptron (MLP) layer applied either in parallel or sequentially, depending on the model architecture.

Given a prompt $\bp \in \R^{k \times V}$, a specific token $t$, layer $\ell$, and that the model applies the MHSA and MLP layers in parallel, the output is given by

\begin{multline}\label{eq:trans-block}
    h_t^{(\ell)} = h_t^{(\ell - 1)} + \mathrm{MHSA}^{(\ell)}(\gamma(h_t^{(\ell - 1)})) \\ +
    \mathrm{MLP}^{(\ell)}(\gamma(h_t^{(\ell - 1)})),
\end{multline}

where $\gamma$ is LayerNorm~\citep{ba2016layernormalization}.

At each layer, we take the last token's activations for both the optimized prompts and the natural language prompts. We use the maximum mean difference (MMD)~\citep{gurnee2023finding} to sort the features in the extracted activations by importance. Specifically, given a set of activations from natural language prompts $\{h^{(\ell)}_{\text{orig},i}\}_{i=1}^N$ and optimized prompts $\{h^{(\ell)}_{\text{optimized},i}\}_{i=1}^M$ at layer $\ell$, the mean difference for feature $j$ is defined as

\begin{align}\label{eq:mean-diff}
    \Delta^{(\ell)}_j = %
    \frac{1}{N}\sum_{i=1}^N h^{(\ell)}_{\text{orig},i,j} - \frac{1}{M}\sum_{i=1}^M h^{(\ell)}_{\text{optim},i,j},
\end{align}

where $h^{(\ell)}_{\cdot,i,j}$ denotes the $j$-th feature (neuron) of the $i$-th example.

We then train a logistic regression classifier using the top identified features with varying levels of sparsity; see Appendix~\ref{app:probe-hparams} for additional training details. Overall, we find that the classifier is able to discriminate ground truth and optimized examples with high accuracy, even with high levels of sparsity; see Figure~\ref{fig:probe-acc}. This is in contrast to the near-random accuracy found in the baseline evaluations which compare original natural language prompts to randomly generated prompts and other natural language prompts. It is important to note that even though for optimized prompts, models clearly contain unique representations which are easily distinguishable from natural prompts, these representations still generate functionally similar outputs.

\subsection{Do optimized prompts rely on a distinct subspace?}

Given that we can classify prompts based on distinct features in the activations, we test the importance of these features in eliciting desired output from optimized prompts. Prior work finds that optimized prompt are sensitive to discrete perturbations~\citep{ishibashi-etal-2023-evaluating,melamed-etal-2024-prompts,cherepanova2024talking}, and we hypothesize that this sensitivity is also present in the model's internal representations. In order to verify our hypothesis, we perform causal intervention and zero-out the top features identified layer-wise via Equation~\ref{eq:mean-diff}, and then measure the top-10 token overlap for each (original, optimized) prompt pair.

Given a prompt $\bp$, let $\P_{\mathrm{LM}}(\cdot|\bp) \in \R^V$ be the output distribution over vocabulary size $V$ at the last position. For layer $\ell$ and hidden state dimension $d$, let $h^{(\ell)} \in \R^d$ be the hidden state output. Define $\mathcal{I}_k \subseteq \{1,\ldots,d\}$ as the indices of the $k$ most important dimensions as found by Equation~\ref{eq:mean-diff}. The intervened distribution is

\begin{align}\label{eq:overlap}
    \P^{(\ell, k)}_{\mathrm{LM}}(\cdot|\bp) = \P_{\mathrm{LM}}(\cdot|\bp; h^{(\ell)}_i = 0 \text{ for } i \in \mathcal{I}_k).
\end{align}

We find that intervening on top features for both natural and optimized prompts has a pronounced effect when compared to the baseline of intervening on random features. However, surprisingly, our experiment contradicts the hypothesis. As shown in Figure~\ref{fig:top-intervened}, for the majority of layers in each model, there is not a large difference between the effect of ablations of top-$k$ features on natural language prompts versus optimized prompts. This means that although optimized prompts may be more sensitive to discrete token-level perturbations than natural language prompts, they do not necessarily share this sensitivity when evaluated from the perspective of the model's internal representations. We do note that for some models, there are specific layers (first and final) which induce a more pronounced change on the output when top features are ablated.

\subsection{How do LMs build predictions from optimized prompts?}\label{sec:layerwise_kl}

Considering both instruction-tuned
and base models, we compute the KL divergence between pairs of optimized and natural prompts at each layer of the model. Specifically, we take the output of the last token $t$ at each layer $\ell$, and multiply it by the final LayerNorm and the LM head in order to project back to the vocabulary space. These outputs are then used to compute the KL divergence between prompt pairs at each layer, and we denote this as $d^{(\ell)}_{KL}(\bp^* || \bp)$.

\begin{figure}[t]
    \centering
    \includegraphics[width=0.47\textwidth]{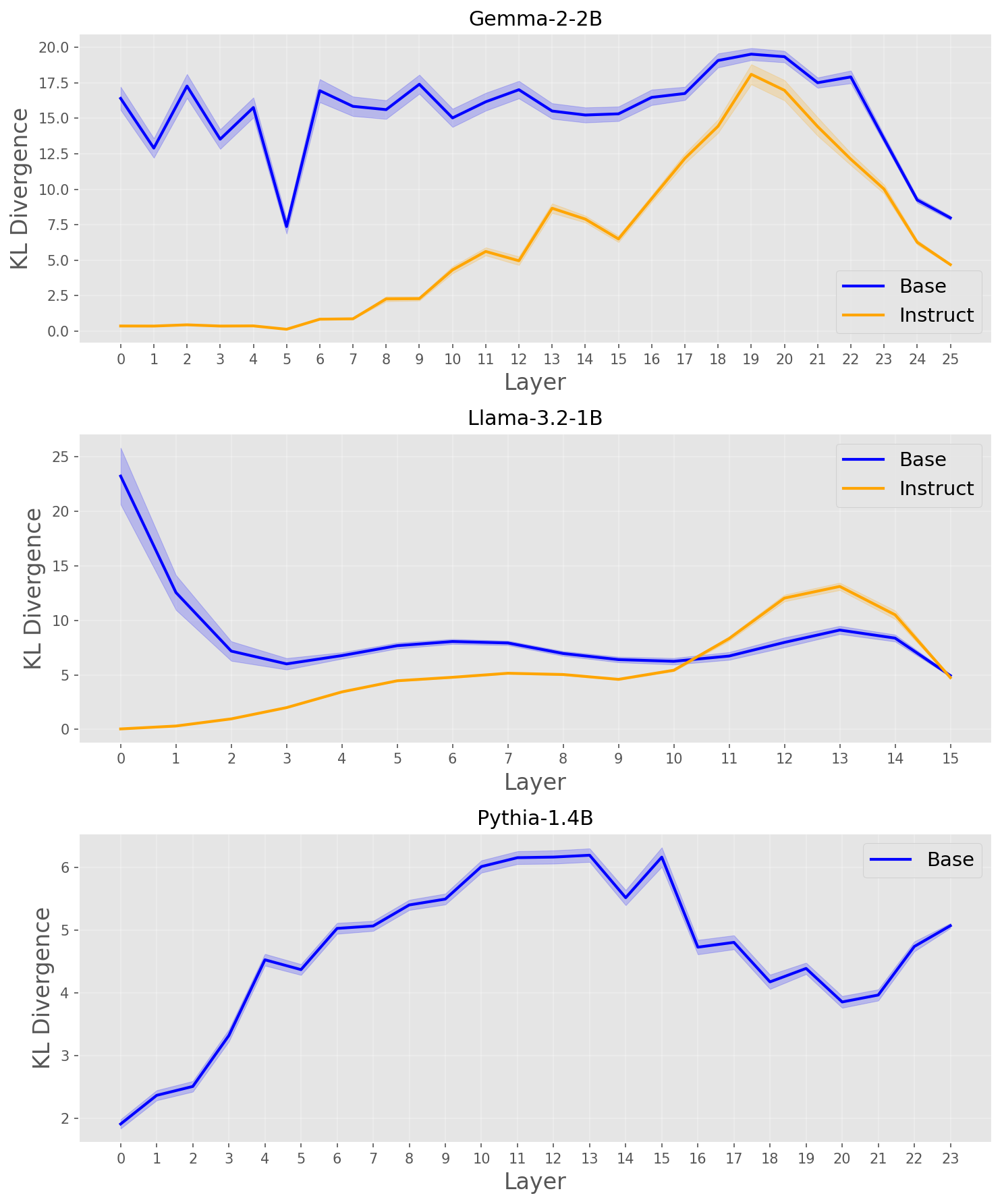}
    \caption{Layer-wise KL divergence. The layer-wise KL divergence is computed as described in Section~\ref{sec:layerwise_kl}. We find that instruction-tuned models follow a similar path; see Figure~\ref{fig:app_layerwise_results_all} in Appendix~\ref{app:full_res} for the full results on all models.
    }
    \label{fig:layerwise-kl}
\end{figure}

Interestingly, as shown in Figure~\ref{fig:layerwise-kl}, across model families, the instruction-tuned versions of the models follow a similar path, with the early layers showing similar representations of prompt pairs, a gradual divergence in the middle layers, and a final sharp trend back to functional similarity in the last few layers. Base models tend to have more divergent representations at the early layers, and also experience a similar sharp trend in the later layers. Clearly, the later layers are crucial for ensuring the functional similarity between the optimized prompts and their natural language counterparts. This supports our findings in Sectio~\ref{sec:probe}, namely that feature ablations in later layers appear to have a stronger effect on optimized prompts, as we can see that the later layers are primarily responsible for aligning the representations of natural language and optimized prompts.

\section{Discussion}

Our work analyzes the mechanisms and ways in which language models parse and interpret discretely optimized prompts. We find that optimized prompts consist primarily of punctuation and noun tokens which are, on average, more rare in the training data than their natural language counterparts. Through sparse probing, we are able to classify optimized prompts and their natural language counterparts with high accuracy. Furthermore, when ablating neurons from model layers, the effectiveness of optimized prompts does not drop in a significant way compared to their natural language counterparts.

One possible application of our analysis is to identify optimized ``jailbreak'' prompts before these prompts are even fully processed by the model. For example, one can train a simple linear classifier on a set of optimized prompts and natural language, and efficiently alert the model provider if a user is inputting suspicious optimized prompts based on the classifier at intermediate layer. Although the majority of prior work has studied optimized prompts through the lens of adversarial attacks, it is also possible that such prompts are benign in nature. It will require further investigation to differentiate benign and malicious optimized prompts.

Finally, prior work finds that discretely optimized prompts transfer between different model families~\citep{rakotonirina2023can,zou2023universaltransferableadversarialattacks,melamed-etal-2024-prompts}. While we focus our discussion on the specific models that generate these prompts, future work can explore how these representations for ``universally transferrable'' optimized prompts differ between models.

\section*{Limitations}

In our work, we primarily consider the ``evil twins'' framework and use GCG~\citep{melamed-etal-2024-prompts, zou2023universaltransferableadversarialattacks}. There are several other techniques for discrete optimization such as FLRT~\citep{thompson2024flrtfluentstudentteacherredteaming} and AutoDAN~\citep{zhu2024autodan} which may induce differing behaviors. Additional work is required to adapt our analysis to these frameworks.

Due to computational constraints, we limit our analysis to models up to 8 billion parameters, as the discrete optimization process requires significant GPU memory and compute. Nevertheless, we test 18 different models. Future work can consider analyzing larger models and investigating trends in optimized prompts across a wider array of model sizes. This will also enable research on scaling effects for prompt optimization.

\bibliography{refs}

\appendix

\section{Additional experimental details}
\label{app:hparams}

\subsection{Tiny stories training setup}\label{app:stories_setup}
We train the word-level Tiny Stories model with a similar model configuration as Pythia-70m~\footnote{\href{https://huggingface.co/EleutherAI/pythia-70m}{https://huggingface.co/EleutherAI/pythia-70m}}. Specifically, we use a batch size of 64, maximum sequence length of 512, hidden dimension of 512, feedforward layer dimension of 2048, 6 layers, and 8 attention heads. For the optimizer and hyperparameters, we choose AdamW~\citep{loshchilov2018decoupled} with $\beta_1 = 0.9$, $\beta_2 = 0.95$, a learning rate of $6 \times 10^{-4}$ with cosine annealing, 500 warmup steps, no gradient accumulation, and 3 epochs of training on a single NVIDIA RTX 6000 Ada GPU.  The word-level tokenizer has a vocabulary size of 46,137 after being trained on the entire Tiny Stories corpus.

\subsection{Optimization and data setup}\label{app:data_setup}

For all models, we perform the evil
twins optimization procedure for 500 steps, with early stopping if $d_{KL}(\bp^* || \bp) \leq 5.0$. In total, we obtain 5000 unique prompts for the word-stories model, and 1200 unique prompts for the open models (from the four aforementioned prompt datasets), with 300 prompts from each dataset. We then filter all final optimized prompts such that $d_{KL}(\bp^* || \bp) \leq 10.0$. Table~\ref{tab:num_filtered_prompts} displays the number of prompts that were optimized for each model after filtering. For the discrete optimization, we use one 8x A100 node.

\begin{figure*}
    \centering
    \includegraphics[width=1.0\linewidth]{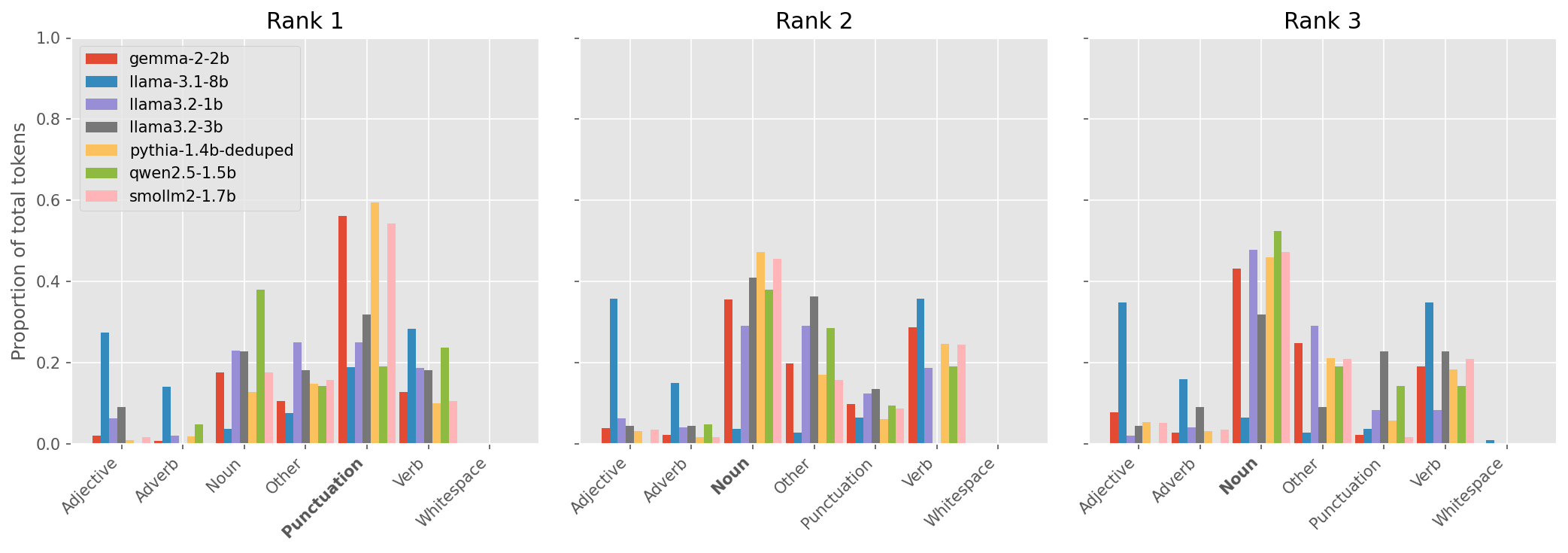}
    \caption{Token category analysis by rank for natural language prompts. For each model and token influence
        rank (as computed in Section~\ref{sec:important-toks}), we show the proportion of tokens belonging to each part of speech category.}
    \label{fig:grammar_graph_orig}
\end{figure*}

\begin{table}[h!]
    \caption{Total optimized prompts after filtering for each tested model.}
    \centering
    \small
    \begin{tabular}{@{}l r@{}}
        \toprule
        \textbf{Model}                                                & \textbf{\# prompts} \\
        \midrule
        gemma-2-2b-base~\citep{gemmateam2024gemmaopenmodelsbased}     & 1156                \\
        gemma-2-2b-instruct~\citep{gemmateam2024gemmaopenmodelsbased} & 1068                \\
        llama3.2-1b-base~\citep{grattafiori2024llama3herdmodels}      & 891                 \\
        llama3.2-1b-instruct~\citep{grattafiori2024llama3herdmodels}  & 765                 \\
        llama3.2-3b-base~\citep{grattafiori2024llama3herdmodels}      & 638                 \\
        llama3.2-3b-instruct~\citep{grattafiori2024llama3herdmodels}  & 624                 \\
        llama3.1-8b-base~\citep{grattafiori2024llama3herdmodels}      & 106                 \\
        pythia-1.4b-base~\citep{biderman2023pythia}                   & 1121                \\
        qwen2.5-0.5b-base~\citep{qwen2025qwen25technicalreport}       & 671                 \\
        qwen2.5-0.5b-instruct~\citep{qwen2025qwen25technicalreport}   & 1081                \\
        qwen2.5-1.5b-base~\citep{qwen2025qwen25technicalreport}       & 532                 \\
        qwen2.5-1.5b-instruct~\citep{qwen2025qwen25technicalreport}   & 1079                \\
        smollm2-1.7b-base~\citep{allal2025smollm2smolgoesbig}         & 887                 \\
        smollm2-1.7b-instruct~\citep{allal2025smollm2smolgoesbig}     & 682                 \\
        smollm2-135m-base~\citep{allal2025smollm2smolgoesbig}         & 1144                \\
        smollm2-135m-instruct~\citep{allal2025smollm2smolgoesbig}     & 1079                \\
        smollm2-360m-base~\citep{allal2025smollm2smolgoesbig}         & 1073                \\
        smollm2-360m-instruct~\citep{allal2025smollm2smolgoesbig}     & 1118                \\
        word-stories                                                  & 2043                \\
        \bottomrule
    \end{tabular}
    \label{tab:num_filtered_prompts}
\end{table}

\subsection{Classifier probe training}\label{app:probe-hparams}

The classifier is a logistic regression trained via scikit-learn~\citep{sklearn_api} for 100 iterations with $l_2$ penalty, \texttt{liblinear} solver, and $1e^{-4}$ convergence tolerance.

\begin{figure}[t]
    \centering
    \includegraphics[width=0.48\textwidth]{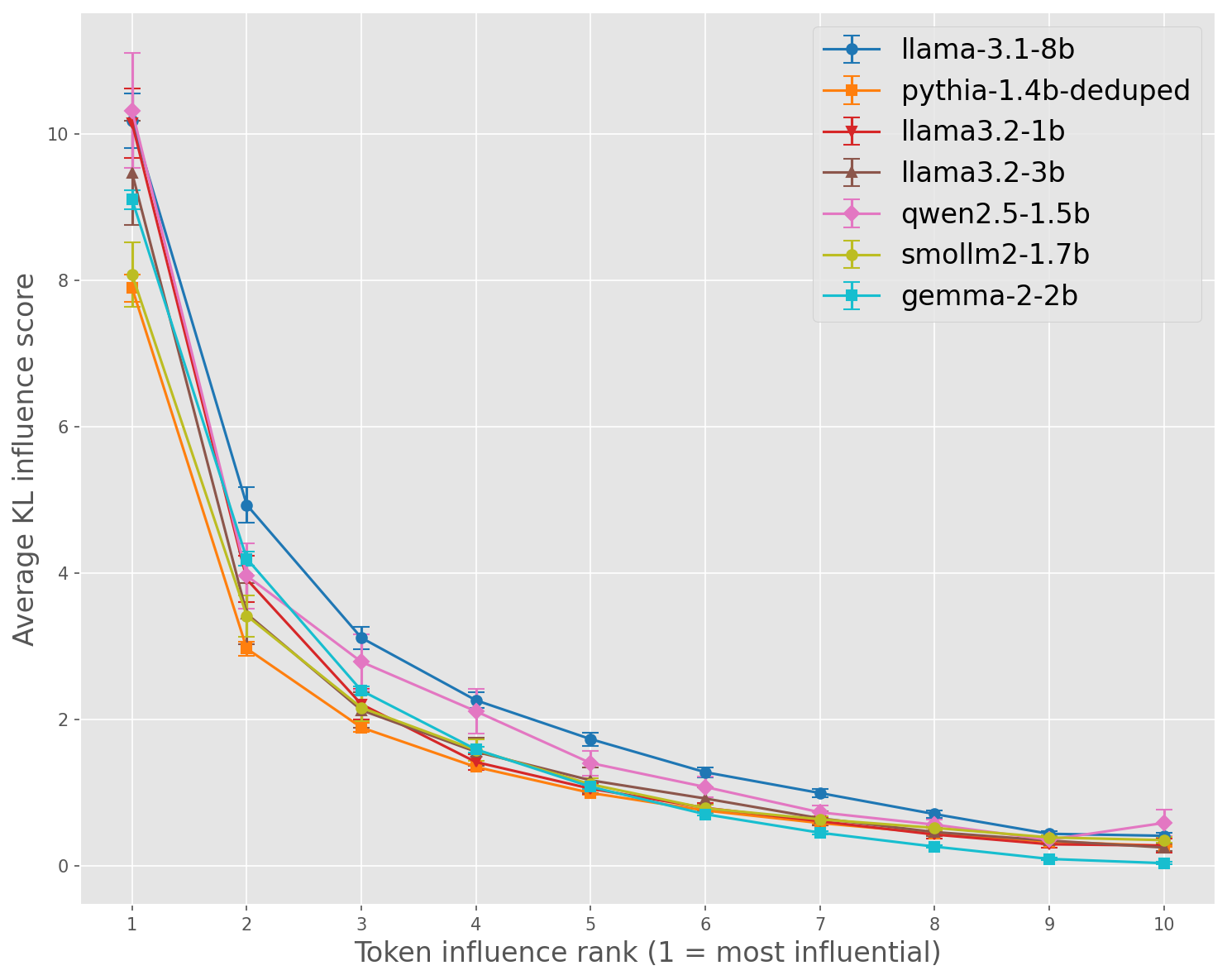}
    \caption{Token rank influence for natural language prompts. The influence score is computed via Equation~\ref{eq:influence_score}.}
    \label{fig:tok_influence_appendix}
\end{figure}

\section{Evil twins and greedy coordinate gradient}
\label{app:gcg}

The Greedy Coordinate Gradient (GCG) algorithm is commonly used to generate adversarial optimized prompts. The procedure starts with an arbitrarily initialized prompt with a fixed number of tokens. At each iteration, it computes the gradient of the loss with respect to each token in the prompt, and identifies some top-$k$ promising replacements for each token in the prompt based on the gradient signal. These candidate replacements for each token are then tested by running a forward pass and taking the new prompt with the lowest loss. We refer the reader to \citealp{zou2023universaltransferableadversarialattacks} for full details regarding the algorithm.

\begin{figure*}[htbp]
    \centering
    \includegraphics[width=0.60\textwidth]{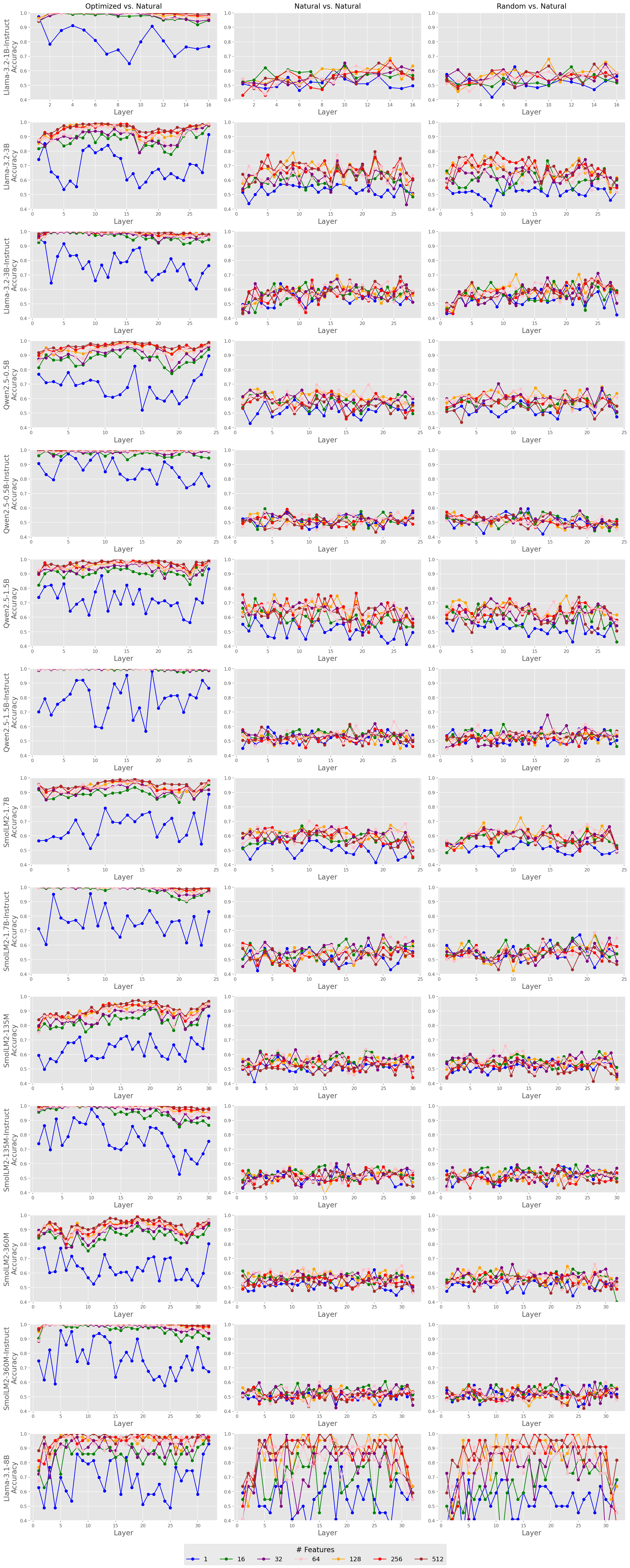}
    \caption{Sparse linear probe results for additional model suits (SmolLM2, Qwen2.5, Llama-3.2)}
    \label{fig:app_probe_results_all}
\end{figure*}

\begin{figure*}[htbp]
    \centering
    \includegraphics[width=0.91\textwidth]{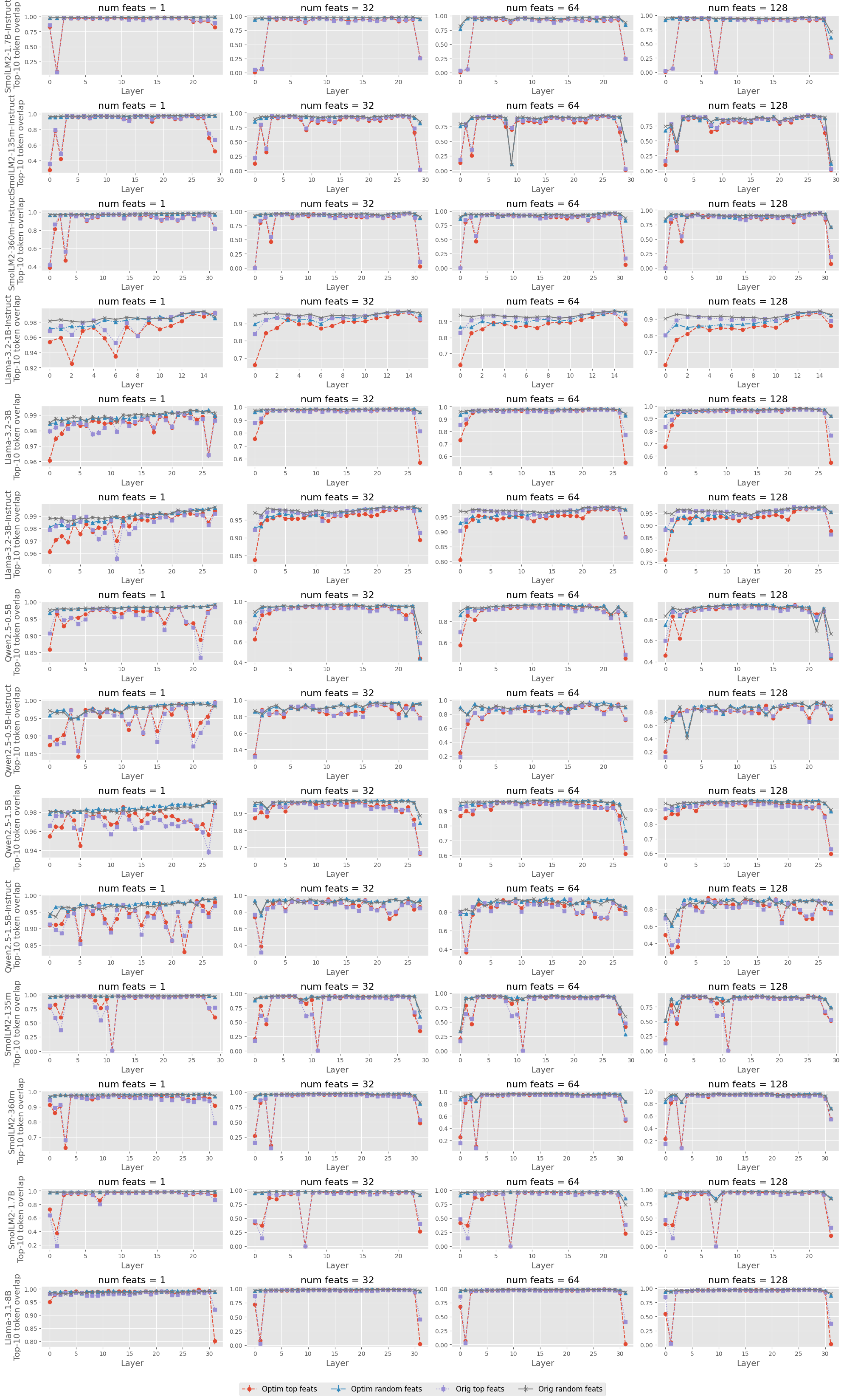}
    \caption{Feature ablation results on all models.}
    \label{fig:app_intervene_results_all}
\end{figure*}

\begin{figure*}[htbp]
    \centering
    \includegraphics[width=0.52\textwidth]{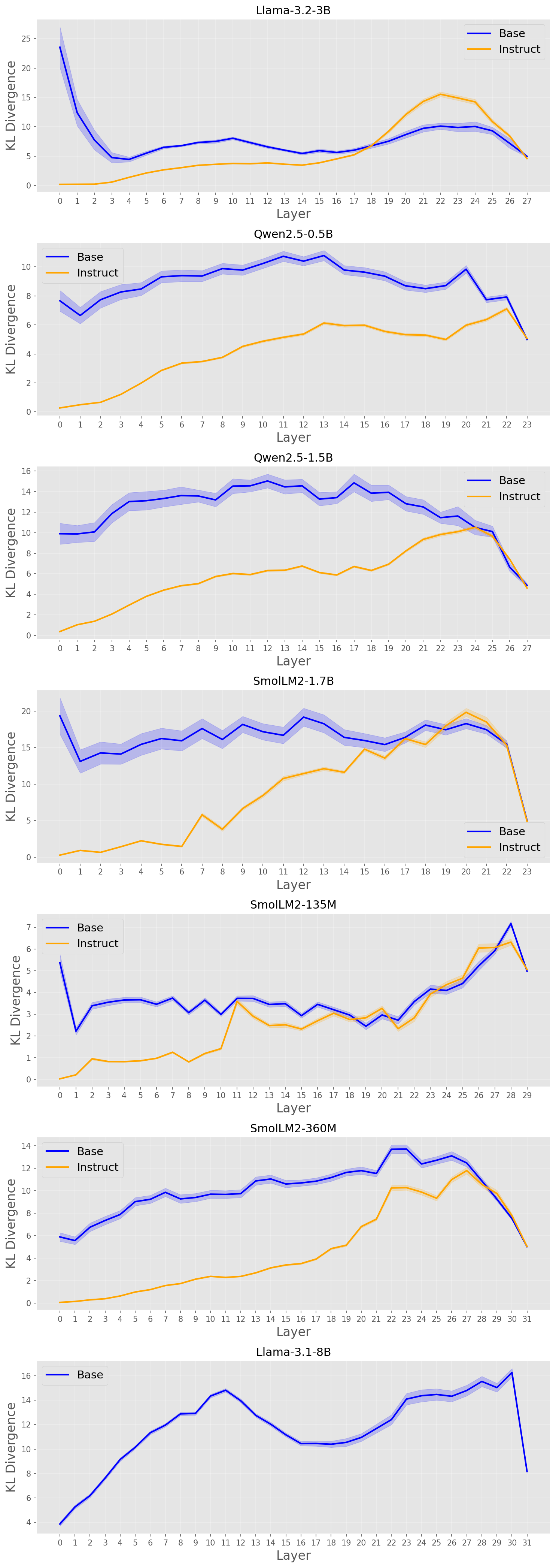}
    \caption{Layer-wise KL Divergence results on all models.}
    \label{fig:app_layerwise_results_all}
\end{figure*}

\section{Composition of natural language prompts}\label{app:orig_removals}

Original natural language prompts have a similar dependency on punctuation and noun tokens as their optimized counterparts; see Figure~\ref{fig:grammar_graph_orig} and Figure~\ref{fig:tok_influence_appendix}.

\section{Experimental results on all models}\label{app:full_res}

We report the full results on the remainder of the 18 tested models for the probing classifier experiment, the feature ablation experiment, and the layer-wise KL divergence experiment. Figure~\ref{fig:app_probe_results_all} displays the results for all model suites on the sparse probing experiment. Figure~\ref{fig:app_intervene_results_all} displays the results for all model suites on the intervention experiment. Figure~\ref{fig:app_layerwise_results_all} displays the results for all model suites on the layer-wise KL divergence experiment.

\end{document}

%% file: preamble.tex
\usepackage{graphicx}
\usepackage{booktabs}
\usepackage{multirow}

\usepackage{float}

\usepackage{amsmath}
\usepackage{amsthm}
\usepackage{amssymb}
\usepackage{color}
\usepackage{mathtools}
\usepackage{makecell}

\usepackage{subcaption}

\providecommand{\R}{\mathbb{R}}
\renewcommand{\P}{\mathbb{P}}

\def\bd{{\boldsymbol d}}

\def\bp{{\boldsymbol p}}

\usepackage{amsmath}
\usepackage{amsthm}
\usepackage{amssymb}
\usepackage{color}
\usepackage{mathtools}
\usepackage{makecell}
\usepackage{todonotes}
\usepackage{diagbox}
\usepackage{tablefootnote}
\usepackage{layouts}

\usepackage{todonotes}

\let\baraccent=\= %
\renewcommand{\=}[1]{\stackrel{#1}{=}} %

\providecommand{\R}{\mathbb{R}}

\renewcommand{\P}{\mathbb{P}}

\mathchardef\mhyphen="2D %

\definecolor{darkblue}{rgb}{0.0, 0.0, 0.55}

\newcommand{\interior}[1]{%
  {\kern0pt#1}^{\mathrm{o}}%
}

\theoremstyle{definition}

\def\bp{{\boldsymbol p}}